\documentclass{article} % For LaTeX2e
\usepackage[preprint]{colm2026_conference}

\usepackage{microtype}
\usepackage{hyperref}
\usepackage{url}
\usepackage{booktabs}

\usepackage{lineno}

\usepackage{enumitem}
\usepackage{amsmath}

\usepackage{graphicx}
\usepackage{adjustbox}

\usepackage{multirow}
\usepackage{booktabs}  
\usepackage{subcaption}

\usepackage{algorithm}
\usepackage{algpseudocode}

\usepackage{verbatim}

\usepackage{xcolor}

\definecolor{darkblue}{rgb}{0, 0, 0.5}
\hypersetup{colorlinks=true, citecolor=darkblue, linkcolor=darkblue, urlcolor=darkblue}

\title{CLSGen: A Dual-Head Fine-Tuning Framework for Joint \\ Probabilistic Classification and Verbalized Explanation}

\author{WonJin Yoon$^{*,1,2}$ \quad
  Kangyu Zhu$^{*,1,\dagger}$ \quad
  Ian Bulovic$^{1}$ \quad
  Autumn Sehy$^{1,\dagger}$ \\
  \textbf{Yanjun Gao$^{3}$ \quad
  Dmitriy Dligach$^{4}$ \quad
  Majid Afshar$^{5}$ \quad
  Timothy A. Miller$^{1,2}$} \\[1em]
  $^{1}$Boston Children's Hospital \quad
  $^{2}$Harvard Medical School \\
  $^{3}$University of Colorado Anschutz Medical Campus \\
  $^{4}$Loyola University Chicago \quad
  $^{5}$University of Wisconsin Madison \\[0.5em]
  {\small $^{*}$Equal contribution. \quad
  $^{\dagger}$Work done at BCH; now at TikTok USA(KZ) and Lightcast (AS).}\\[0.5em]
    \texttt{\{firstname.lastname\}@childrens.harvard.edu} \\
\texttt{yanjun.gao@cuanschutz.edu}, \texttt{ddligach@luc.edu}, \texttt{mafshar@medicine.wisc.edu} \\
}

\begin{document}

\ifcolmsubmission
\linenumbers
\fi

\maketitle

\begin{abstract}
With the recent progress of Large Language Models (LLMs), there is a growing interest in applying these models to solve complex and challenging problems. Modern LLMs, capable of processing long contexts and generating verbalized explanations, offer significant potential in addressing real-world applications.
However, a critical hurdle in deploying LLMs for practical decision-making is their inability to provide reliable, quantitative probabilities. While task-specific fine-tuning of LLMs using traditional discriminative objectives (similar to encoder-only models) can yield probability estimates, this often leads to catastrophic forgetting and linguistic collapse. Consequently, the model loses its ability to generate explanations, severely undermining its interpretability and usability.
To address this challenge, we propose CLSGen, a novel LLM fine-tuning framework designed for binary classification tasks. The CLSGen framework encompasses a new model architecture, training methodology, and data construction strategy to enable robust probability estimation without sacrificing the model's inherent explanation-generation capabilities.
Experimental results across multiple benchmark datasets demonstrate that models fine-tuned with CLSGen outperform existing baselines in classification metrics (AUROC and F1-score). Regarding explanation, the results showed strong alignment between predicted labels and generated justifications, as well as high readability.
\end{abstract}

\section{Introduction}

Recently, Large Language Models (LLMs) have demonstrated progress in performance in complex reasoning tasks including those requiring lengthy inputs~\citep{wang2024beyond}, opening the opportunity for expansion of their application into specialized domains such as clinical and biomedical fields. In particular, applications in patient outcome prediction and drug discovery have garnered significant attention, highlighting the immense potential of LLMs in these high-stakes areas~\citep{thirunavukarasu2023large, benshoham2024cpllm}. In response to this trend, domain-specific benchmarks such as LCD benchmark \citep{yoon2025lcd} and CliniFact \citep{zhang2025clinifact} have been proposed.

However, several hurdles remain before these models can be effectively deployed in real-world scenarios. In fields requiring high-stakes decision-making, users demand not only high performance but also (1) Confidence Estimation (how certain the model is) and (2) Explanation (why the model made such a decision)~\citep{begoli2019need, zhao2024explainability}. While modern LLMs have become increasingly capable of providing plausible verbalized explanations, quantifying internal confidence or probability remains an area of active research. In this study, we propose a novel fine-tuning framework, within the specific context of binary classification, that provides both probability and explanation.

While various attempts have been made to derive probabilities from LLMs without fine-tuning, each possesses inherent limitations (detailed in the Related Work section). A straightforward approach to producing probabilities, which is well-established in encoder-only model research, is to utilize the model for contextual representation by adding a Sigmoid layer to the final output and fine-tuning with a Binary Cross-Entropy (BCE) loss. However, this method can lead to catastrophic forgetting ~\citep{mccloskey1989catastrophic, luo2025empirical}, where the model loses its original capabilities. In this case, text generation, which constitutes a core advantage of LLMs, can be lost. Consequently, the model becomes unable to provide the verbalized explanations (More details in Section~\ref{sec:motiv}).

To overcome these limitations, we propose a comprehensive fine-tuning framework, \textit{CLSGen}, that consists of a new model architecture, training methodology, and data construction strategy. Our framework incorporates both classification and data generation into the training objective to ensure the text generation quality is maintained, allowing the model to provide both classification probability and verbalized explanations. The proposed architecture utilizes a decoder-only model equipped with a dual-head structure: a standard language modeling (LM) head for text generation and an additional Sigmoid layer for classification. Regarding data construction, we employ a strategy inspired by weakly supervised learning, utilizing an LLM-based method to generate high-quality target text pairs as training data.

We validated our framework on challenging, real-world scenario benchmarks, including datasets characterized by long-context requirements that existing models often struggle to solve. Our models were evaluated through a diverse range of metrics, including classical measures (AUROC, F1-score), statistical measures (Cohen’s Kappa), and qualitative assessments via LLM-as-a-judge. The results demonstrate that our approach outperforms inference-only and baseline fine-tuning methods in classification performance while achieving high scores in both readability and consistency between classification and generation outputs. 

\section{Related Work}

\subsection{Probability estimation for LLMs}

%\Done{By KZ} 
Existing approaches to probability estimation for LLMs can be broadly categorized into single-run and iteration-based methods. Single-run methods derive a confidence score from a single model inference. One instance of this type of method, token probability approaches, extracts confidence from the model's internal log-probabilities over the predicted label tokens, leveraging the model's intrinsic calibration~\citep{kadavath2022language, gu2024probabilistic}. Verbalized probability methods, another single-run type of method, instead prompt the model to explicitly output a numerical confidence score as part of its response~\citep{xiong2023can, yang2024verbalized}. While computationally efficient, both approaches carry known limitations: token probabilities are sensitive to tokenization choices and do not reflect the full reasoning process, whereas verbalized confidence is prone to systematic overconfidence and miscalibration~\citep{xiong2023can, yang2024verbalized, gao2025uncertainty, heo2025llms}. 

Iteration-based methods run the model multiple times to produce a more robust probability estimate. Self-consistency~\citep{wang2023selfconsistency} samples diverse reasoning paths and uses majority voting as a confidence proxy. Semantic entropy~\citep{kuhn2023semantic} takes a complementary approach by clustering semantically equivalent generations and computing uncertainty over meaning rather than surface form. Tournament-based approaches extend this by performing pairwise comparisons across multiple runs, demonstrating strong discrimination performance~\citep{yoon2025tournament,shrivastava2025language}; however, multi-label settings remain largely unexplored. A shared limitation across all iteration-based methods is their high computational cost; moreover, they yield a scalar confidence value without any accompanying verbal explanation, which is critical for high-stakes decision-making.

\subsection{Reinforcement Learning based finetuning}
Reinforcement learning (RL)-based finetuning represents another paradigm for aligning LLM outputs with desired behaviors. Reinforcement Learning from Human Feedback (RLHF)~\citep{ouyang2022training} trains a reward model from human preference annotations and optimizes the language model policy via PPO. Direct Preference Optimization (DPO)~\citep{rafailov2023direct} simplifies this pipeline by directly optimizing on preference pairs without a separate reward model. More recently, Group Relative Policy Optimization (GRPO)~\citep{shao2024deepseekmath} eliminates the need for a critic model by using group-level reward normalization, and outcome-reward RL has demonstrated strong results in training reasoning-capable models~\citep{guo2025deepseekr1}. While our approach shares the goal of aligning outputs with desired properties, it differs fundamentally: rather than optimizing a reward signal, we jointly fine-tune classification and generation objectives through a dual-head architecture, avoiding the instability and complexity of RL-based training.

\section{Motivation}\label{sec:motiv}

Self-attention is the mechanism for computing sequence representations \citep{vaswani2017attention}. In the early stages of Transformer research, encoder-only models such as BERT predominantly utilized the output of a specific position, typically the first token, [CLS], as a global representation for classification tasks. Similar approaches to provide vectorized representations of tokens or sequences have recently been proposed for decoder-only models as well (e.g., Gemini Embedding \citep{lee2025gemini}). However, since LLMs are pre-trained for text generation, fine-tuning a decoder-only LLM carries a significant risk of catastrophic forgetting or linguistic collapse, wherein the model loses its previously learned knowledge about language. This can severely degrade text generation performance, which is the core strength of these models.

Before presenting our proposed framework, we present preliminary experimental results to empirically validate whether such catastrophic forgetting actually occurs in large-scale LLMs. Proving that generative capabilities are destroyed in the process of simple classification fine-tuning provides strong justification for a new architecture that can both provide classification probability and preserve language capability.

For this experiment, we provided the model with prompts requiring responses in a specific structured format and calculated the \textit{Number of parsable outputs} by verifying if the generated results could be parsed according to predefined rules. (Detailed information is available in Appendix A.1). This metric serves as a direct measure of instruction-following capability and acts as a robust proxy for overall text generation quality.

\begin{figure}[t]
\centering
\begin{adjustbox}{center}
\begin{minipage}{\linewidth}
    \centering
    \begin{subfigure}{0.328\linewidth}
        \centering
        \includegraphics[width=\linewidth]{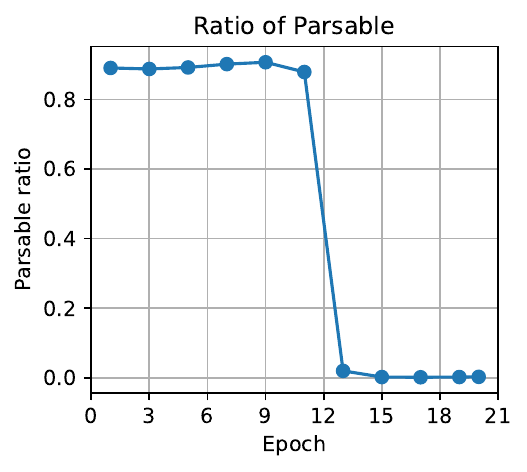}
        \caption{Parsability}
        \label{fig:parsable}
    \end{subfigure}
    \begin{subfigure}{0.331\linewidth}
        \centering
        \includegraphics[width=\linewidth]{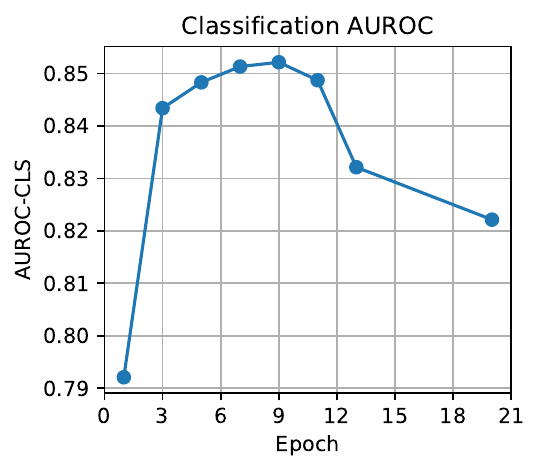}
        \caption{Classification AUROC}
        \label{fig:auroc}
    \end{subfigure}
    \begin{subfigure}{0.328\linewidth}
        \centering
        \includegraphics[width=\linewidth]{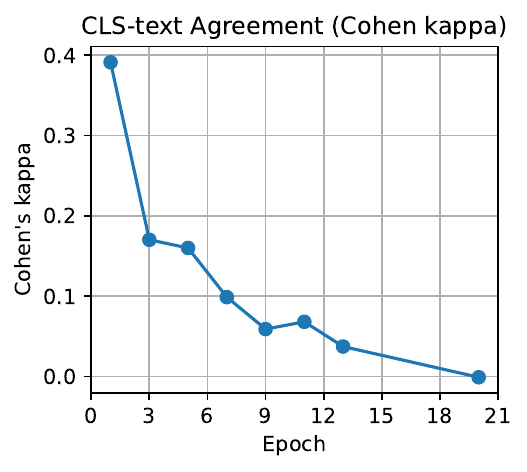}
        \caption{Text and class agreement}
        \label{fig:kappa}
    \end{subfigure}
\end{minipage}
\end{adjustbox}

\caption{Classification and generation quality during classification-only training.}
\label{fig:clsonly-Parsable}
\end{figure}

Figure~\ref{fig:clsonly-Parsable} (a) shows the experimental results showing the phenomenon of linguistic collapse. 
After the 11th epoch, the model's generative capability virtually vanishes. The following are two representative examples from the 13th epoch. 

\begin{itemize}[itemsep=0.5em, topsep=0pt, leftmargin=2em, label=\textbullet]
    \item Pom Pomuppy Pom Pom PoméricRY Pom Pom Pom Pom waveformuppybagriprip
    \item omicçãourygação terminallyuppyuppyção terminallyuppyuppyção terminvidação
\end{itemize}

Most other samples exhibit a similar pattern of linguistic collapse, where the model outputs meaningless, contextless text. These results demonstrate that optimizing solely for classification leads to a loss of the LLM's intrinsic advantage: text generation. This underscores the necessity of our CLSGen framework, which ensures classification accuracy while maintaining the ability to provide verbalized explanations.

In addition to the parsable outputs measure, we evaluated classification-text agreement using Cohen’s kappa. This metric quantifies the extent to which generated text aligns with classification outputs. Figures~\ref{fig:clsonly-Parsable} (b) and (c) summarize the transition of classification accuracy and the concurrent degradation of text generation quality as fine-tuning progresses. The degradation of generation performance begins at the onset of classification training; the alignment score (Cohen’s kappa, threshold=0.5) declines from the start of training, indicating that generated text becomes progressively less consistent with classification outcomes. These results show that although linguistic collapse does not immediately occur, the degradation of text quality and misalignment begins as soon as training starts.

\section{Method}

Our approach consists of two phases: \textit{data construction} and \textit{CLSGen training} (Figure~\ref{fig:main}). We first define the task, then describe data construction and the CLSGen training method.

\subsection{Task definition}
Given a dataset $\mathcal{D} = \{(x_i, y_i)\}_{i=1}^N$, each instance pairs an input token sequence $x_i = (t_{i,1}^x, t_{i,2}^x, \dots, t_{i,n}^x)$, and a target label $y_i \in \{0, 1\}$. In the \textsc{LCD} benchmark, $x_i$ represents a discharge note and $y_i$ represents 30-day or 90-day mortality. For CliniFact, $x_i$ is a pair consisting of a scientific claim and an abstract of clinical literature, while $y_i$ denotes whether the literature supports the claim.
The data construction step generates an augmented dataset $\mathcal{T} = \{(x_i, y_i, e_i)\}$, where each explanation $e_i = (t_{i,1}^e, t_{i,2}^e, \dots, t_{i,m}^e)$ is generated by a large language model (LLM) $\mathcal{M}_{\text{data}}$ and provides a rationale connecting $x_i$ to $y_i$. Finally, the CLSGen training step fine-tunes a base LLM $\mathcal{M}_{\text{base}}$ on $\mathcal{T}$ to produce the target model $\mathcal{M}_{\text{CLSGen}}$.

\subsection{Data construction}

\begin{algorithm}[b]
\small
\caption{Explanation-Augmented Data Generation}
\label{alg:data_gen}
\begin{algorithmic}[1]
\State \textbf{Input:} Dataset $\mathcal{D} = \{(x_i, y_i)\}_{i=1}^N$, Language model $\mathcal{M}_d$, Max trials $K$
\State \textbf{Output:} Augmented dataset $\mathcal{T} = \{(x_i, y_i, e_i)\}$ 

\State $\mathcal{T} \leftarrow \emptyset$
\For{each $(x_i, y_i) \in \mathcal{D}$}
    \For{$k = 1$ \textbf{to} $K$}
        \State $o_i^{(k)} \leftarrow \text{Generate}(\mathcal{M}_d, x_i)$ \Comment{Generate explanation and prediction}
        \State $(\hat{y}_i^{(k)}, e_i^{(k)}) \leftarrow \text{Parse}(o_i^{(k)})$ \Comment{$\hat{y}_i^{(k)}$ is parsed from verbalized label}
        \If{$\text{isValid}(e_i^{(k)})$ \textbf{and} $\hat{y}_i^{(k)} = y_i$}
            \State $\mathcal{T} \leftarrow \mathcal{T} \cup \{(x_i, y_i, e_i^{(k)})\}$
            \State \textbf{break} \Comment{Store first correct instance and move to next $i$}
        \EndIf
    \EndFor
\EndFor
\end{algorithmic}
\end{algorithm}
This step, formalized in Algorithm~\ref{alg:data_gen}, aims to ensure both the quality of the explanations and their alignment with the ground-truth labels of the training dataset for CLSGen training.
For each input $x_i$, we prompt $\mathcal{M}_{\text{data}}$ to generate a predicted label and an explanation in a specified format. We sample $k$ outputs from the model and retain the first explanation $e_i$ that both adheres to the format and produces a prediction $\hat{y} = y_i$. Samples failing to meet both criteria across all $k$ attempts are discarded from the resulting dataset $\mathcal{T}$.

\subsection{CLSGen training}

\begin{figure}[h]
  \centering
  \includegraphics[width=\textwidth]{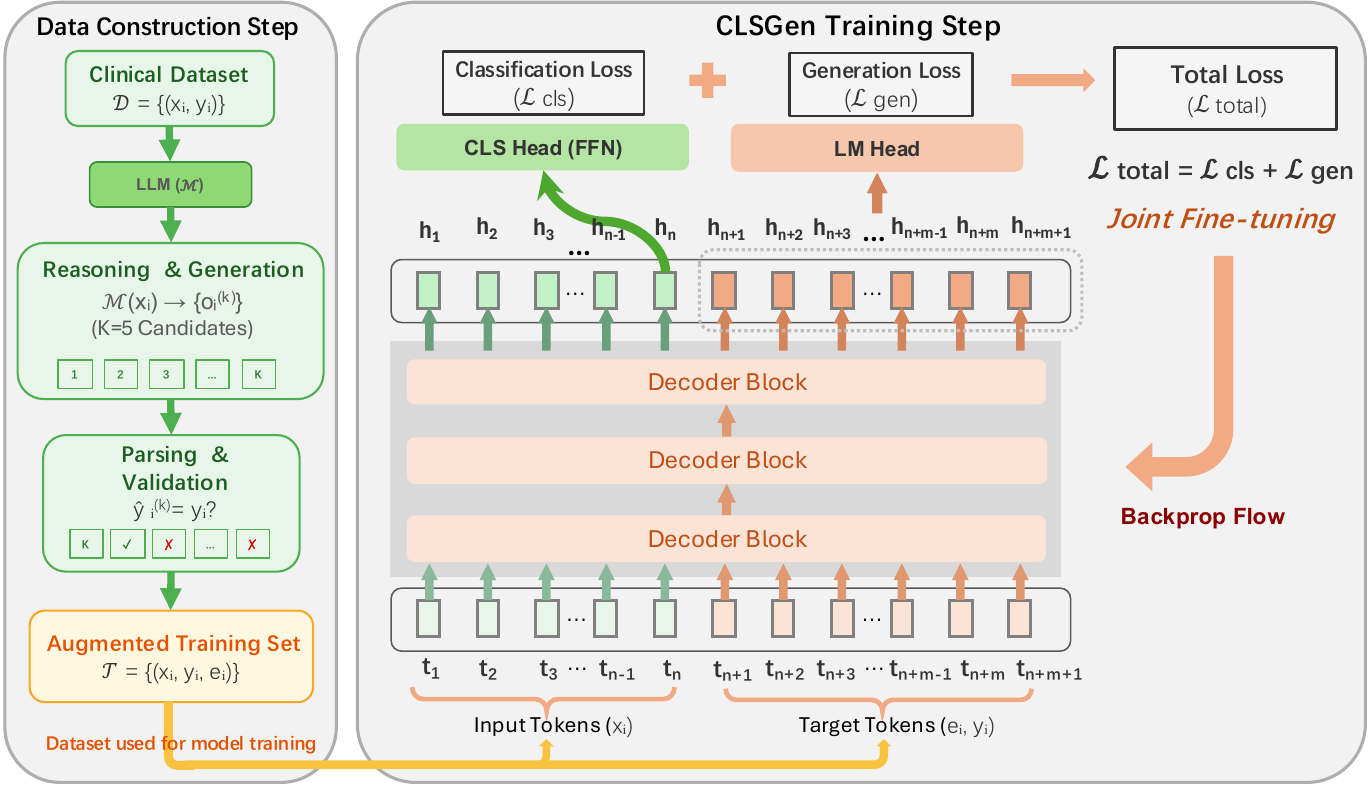}
  \caption{Diagram of data construction pipeline (left) and CLSGen fine-tuning (right).}
  \label{fig:main}
\end{figure}

We propose a jointly trained model that combines rationale generation (i.e., explanation) with binary classification over a shared pretrained language model backbone. Given an input document, the model is prompted with a structured template that includes an instruction, the input document, and a generation prefix (e.g., \textit{Reasoning:}), which together form the input sequence. The last token of  input sequence is denoted as $t_n$ in the right panel of figure~\ref{fig:main}. 

The model is trained under two complementary loss terms.
First, a causal language modeling objective $\mathcal{L}_{\text{gen}}$ trains the model to predict the target sequence consisting of the explanation $e_i$ followed by a structured classification token and the verbalized label $y_i$ (e.g., \textit{Classification: 1:death}). While optimized via teacher forcing during training, this objective enables the model to autoregressively generate a textual rationale and classification at inference time.
Second, a classification objective $\mathcal{L}_{\text{cls}}$ extracts the hidden representation $\mathbf{h}_n$ corresponding to the last token of the input prefix, $t_n$. This vector is passed through a two-layer MLP classification head to predict the binary label directly. The total training objective minimizes the sum of both loss terms:
$$
\mathcal{L}_{\text{total}} = \mathcal{L}_{\text{gen}} +  \mathcal{L}_{\text{cls}}
$$
$$
 \mathcal{L}_{\text{cls}} = \text{CrossEntropy}\left(\text{MLP}\left(\mathbf{h}_n\right),\ y\right)
$$

We adapt the attention projections using Low-Rank Adaptation (LoRA; \citet{hu2022lora}). Both the LoRA parameters and the classification (CLS) head are optimized end-to-end, allowing gradients from both objectives to jointly shape the shared representations over the full input context.

\section{Experimental Setup}

\subsection{Datasets}
We use three tasks from two datasets in our experiments. Detailed descriptions are provided below, and dataset statistics are summarized in Table~\ref{tab:dataset_stats}.

\begin{table}[h]
\small
\centering
\begin{tabular}{lrrrr}
\toprule
\textbf{Dataset} & \textbf{Train} & \textbf{Dev.} & \textbf{Test} & \textbf{Total} \\
\midrule
LCD benchmark & 34{,}759 & 7{,}505 & 7{,}568 & 49{,}832 \\
CliniFact     & 1{,}260  & 316     & 394     & 1{,}970  \\
\bottomrule
\end{tabular}
\caption{Dataset statistics. Both 30 and 90 days tasks of the LCD benchmark have identical splits.}
\label{tab:dataset_stats}
\end{table}

\subsubsection{LCD benchmark} 
The LCD benchmark \citep{yoon2025lcd} is a binary classification task that predicts patient prognostic outcomes after discharge using discharge notes. The main task is to determine whether a patient will survive within 30 days of discharge. In addition, the dataset includes auxiliary prediction tasks for longer time horizons (60 and 90 days). The textual data are derived from MIMIC-IV and MIMIC-IV-Note \citep{johnson2023mimic}, where the date of death information is supplemented with Massachusetts state vital records of death certificates to capture out-of-hospital deaths. The dataset is publicly accessible under a Data Use Agreement, which requires completion of training in human subjects research protections and compliance with HIPAA regulations.

% We note that the LCD benchmark is highly challenging for several reasons. The text length is very long, requiring models to reason and synthesize over long distances. In addition, the prevalence is quite low, as the benchmark explicitly excludes the cases with discharge to hospice care that would be easiest to predict. But most significantly, it is inherently challenging to make a prediction about the future -- unlike many clinical classification tasks which amount to information finding, information about future survival is not present in discharge summaries. There is inherent uncertanty in this kind of question, and so we believe the ceiling on performance is quite low.
The LCD benchmark is particularly challenging: inputs are long, class prevalence is low (hospice cases are excluded), and predicting future mortality is inherently uncertain since discharge summaries contain no direct signal about post-discharge survival.

\subsubsection{CliniFact}
%\Done{By KZ} 
CliniFact \citep{zhang2025clinifact} is a binary classification dataset for evaluating clinical research claim verification. Each instance consists of a scientific claim derived from a clinical trial protocol paired with the corresponding PubMed abstract, and the task is to determine whether the abstract provides positive support for the claim (\textit{TRUE}) or not (\textit{FALSE}). The dataset contains 1,970 instances drawn from 992 unique clinical trials spanning 22 disease categories and linked to 1,540 unique publications. It is split into train, validation, and test sets, with label distribution reported across all splits.

Unlike LCD, CliniFact requires scientific reasoning over claim-evidence pairs, serving as a complementary benchmark for evaluating generalizability beyond clinical narrative text.

\subsection{Baselines}

We compare our proposed CLSGen models with three inference-time prompting settings for large language models (LLMs) as well as a traditional Convolutional Neural Networks (CNN) baseline. The LLM-based settings do not involve additional training and differ only in their output prompting strategies: \textit{verbalized probability}, \textit{label prediction}, and \textit{self-consistency}.

\textbf{Verbalized probability} prompts the model to explicitly output a probability estimate (e.g., \textit{"What is the probability (0 to 100)?"} and \textit{"State your answer in the format: Probability: [integer]"}).

\textbf{Label prediction} instructs the model to produce a discrete class label in a fixed format (e.g., \textit{"Answer with exactly one of: 0:alive or 1:death."}).

\textbf{Self-consistency} \cite{wang2023selfconsistency} is implemented as majority voting over 10 independent runs using the label prediction setting.

The \textbf{CNN} is trained from scratch in a supervised manner; detailed hyperparameters are provided in the appendix.

\subsection{Evaluation metrics}

\subsubsection{Classification metrics}
We employed two primary categories of metrics to evaluate classification performance. Since our primary objective is to achieve precise probability estimation, we designated Area Under the Receiver Operating Characteristic Curve (AUROC) as our primary metric. AUROC is a threshold-independent metric that evaluates the raw probabilities produced by the model without artificial binarization. It essentially measures the ranking quality: an AUROC of 1.0 signifies that every positive sample is assigned a higher probability than any negative sample, whereas an AUROC of 0.5 indicates that the model's predictions are near random, possessing no discriminatory power between classes. 

Since probabilities could not be extracted from verbalized prediction experiments, we utilized Precision, Recall, and F1-score as alternative metrics. Unlike AUROC, these metrics are threshold-dependent as they rely on binarized predictions. The F1-score is particularly sensitive to the threshold because it is the harmonic mean of precision and recall, which penalizes imbalances between the two. To account for this sensitivity, we report results under two settings: (1) a default threshold of 0.5, and (2) an F1-optimized threshold found using the development set.

\subsubsection{CLS-Text consistency metrics}
We also evaluate the alignment between the classification head’s output and the verbalized text output. AUROC-Alignment and Cohen’s Kappa measure the correspondence between probabilities and textualized labels (e.g., [0:alive, 1:death]), playing a key role in interpreting our experimental results. Furthermore, we employ LLM-as-a-judge to assess whether the generated explanations logically align with the predicted labels in the generated text.

\textbf{AUROC-Alignment:} AUROC-Alignment is a metric proposed in this study, extending the standard AUROC. The conventional AUROC takes a continuous random variable $\mathbf{X} \in (0,1)^n$ and a set of binary gold labels  $\mathbf {Y} \in \{0,1\}^n$ (where $n$ is the number of test samples), to compute $Score_{AUROC} = \text{AUROC}(X, Y)$. We define AUROC-Alignment by substituting the gold labels $\mathbf{Y}$ with the verbalized labels $\mathbf{\hat{Y}}$: $Score_{AUROC-alignment} = \text{AUROC}(X, \hat{Y})$
Here, $X$ represents the probability derived from the classification head, and $\hat{Y}$ denotes the verbalized binary label extracted from the generated text. This metric quantifies the degree to which the model's classification head output is closely aligned with its textual output.

\textbf{Cohen’s kappa ($\kappa$):} Cohen’s kappa evaluates the consistency between two categorical label assignments while correcting for random agreement. In this study, we binarize the probability outputs from the classification head using a threshold to obtain one set of predictions, and treat the verbalized label in the generated text as another. Agreement between these two sets is measured using Cohen’s kappa.

The kappa score ranges from -1 to 1, where 1 indicates perfect agreement and 0 corresponds to chance-level agreement. Values 0.01--0.20 indicate slight agreement; 0.21--0.40 fair; 0.41--0.60 moderate; 0.61--0.80 substantial; and 0.81--1.00 almost perfect agreement \citep{landis1977measurement}. 

\textbf{LLM-as-a-Judge:} We evaluate the internal consistency between the generated explanation ($e$) and the verbalized label ($\bar{y}$) using an LLM-as-a-judge approach. Specifically, we prompt an external LLM to infer the label ($\hat{y}_{inf}$) implied by the generated rationale $e$, without providing the model's actual predicted label. We then compare this inferred label to the model's verbalized prediction $\bar{y}$. We define two metrics for this alignment: \vspace{0.5em}\\
\textbf{Rationale-Label Inconsistency (RLI):} The proportion of samples where the inferred label does not match the verbalized label ($\hat{y}_{inf} \neq \bar{y}$), calculated over successfully parsed outputs. \\
\textbf{Rationale-Label Kappa ($\kappa_{RL}$):} The agreement (Cohen’s Kappa) between $\bar{y}$ and $\hat{y}_{inf}$. 

Additionally, we assess the readability of the generated output to ensure clinical utility. Prompts are provided in Appendix~\ref{appendix:sec:prompt}.

\subsection{Implementation Details}

We use \texttt{Llama-3.1-8B-Instruct} \citep{grattafiori2024llama} as $\mathcal{M}_{\text{base}}$ for fine-tuning and zero-shot baselines. We fine-tune on four NVIDIA A100 (80GB) GPUs for about five days  for LCD benchmark and four hours for CliniFact. We tune the number of training epochs and the classification threshold on the development set; all remaining implementation details including hyperparameter values, selection criteria, prompt, and details about data construction are reported in the reproducibility statement and Appendix~\ref{appendix:sec:details}. 

Reported values for label prediction, verbalized probability, and CNN are averaged over 10 independent runs. For self-consistency, we use majority voting across runs, excluding unparseable outputs on a per-sample basis.

\begin{table*}[t]
\small
\centering
\resizebox{\columnwidth}{!}{
\begin{tabular}{llcccccc}
\toprule
 &  & \multicolumn{3}{c}{Inference-only methods} & Traditional & \multicolumn{2}{c}{\textbf{CLSGen (Ours)}} \\
\cmidrule(lr){3-5} \cmidrule(lr){6-6} \cmidrule(lr){7-8}
Dataset/task & Metric & Verb prob & Label pred & Self cons & CNN & Default & Threshold \\
\midrule

\multirow{5}{*}{\shortstack{LCD benchmark \\ 30 days}}
& \textbf{AUROC}  & 0.6598 & N/A    & 0.7455 & 0.8582 & \multicolumn{2}{c}{0.8586} \\
& \textbf{F1}     & 0.1409 & 0.2023 & 0.2853 & 0.2182 & 0.2263 & 0.2455 \\
& Prec.           & 0.0809 & 0.1443 & 0.2315 & 0.5889 & 0.5606 & 0.5616 \\
& Recall          & 0.5467 & 0.3383 & 0.3716 & 0.1364 & 0.1418 & 0.1571 \\
& Parsable        & 0.8098 & 0.8660 & 0.9432 & N/A & \multicolumn{2}{c}{0.9991} \\

\midrule

\multirow{5}{*}{\shortstack{LCD benchmark \\ 90 days}}
& \textbf{AUROC} & 0.6255 & N/A    & 0.7164 & 0.8360 & \multicolumn{2}{c}{0.8574} \\
& \textbf{F1}    & 0.2379 & 0.2623 & 0.3070 & 0.3108 & 0.3224 & 0.3610 \\
& Prec.          & 0.1578 & 0.2195 & 0.2878 & 0.5991 & 0.5808 & 0.4818 \\
& Recall         & 0.4831 & 0.3262 & 0.3289 & 0.2101 & 0.2232 & 0.2886 \\
& Parsable       & 0.7956 & 0.8548 & 0.9442 & N/A & \multicolumn{2}{c}{0.9762} \\

\midrule

\multirow{5}{*}{CliniFact}
& \textbf{AUROC} & 0.9032 & N/A    & 0.9383 & 0.8527 & \multicolumn{2}{c}{0.9457} \\
& \textbf{F1}    & 0.7844 & 0.8181 & 0.8559 & 0.6212 & 0.8597 & 0.8532 \\
& Prec.          & 0.7762 & 0.8413 & 0.8716 & 0.6567 & 0.8796 & 0.8857  \\
& Recall         & 0.7938 & 0.7965 & 0.8407 & 0.5912 & 0.8407 & 0.8230  \\
& Parsable       & 0.9888 & 0.9832 & 1.0000 & N/A & \multicolumn{2}{c}{1.0000} \\

\bottomrule
\end{tabular}
}
\caption{Performance comparison across datasets and methods. \textit{Default} uses a fixed threshold of 0.5, while \textit{Threshold} uses thresholds tuned on the development set. Standard deviations are provided in Appendix~\ref{appendix:sec:stats} (Table~\ref{tab:appendix-std}).
}\label{tab:main-perf}
\end{table*}

\section{Results}

Table~\ref{tab:main-perf} summarizes the predictive performance of our proposed CLSGen compared to baselines. The corresponding standard deviations across 10 independent runs are reported in Appendix~\ref{appendix:sec:stats} (Table~\ref{tab:appendix-std}).
In most experimental settings, CLSGen outperforms both inference-only methods and the traditional CNN across key metrics, specifically AUROC and F1-score.
Notably, while Self-consistency shows a higher F1 on the 30-day LCD benchmark task, given that CLSGen maintains a substantially higher AUROC in the same task, this discrepancy is likely attributable to the threshold-dependent nature of the F1 metric rather than a lack of fundamental discriminative power. In other words, while CLSGen demonstrates superior probabilistic potential, its F1-score may vary depending on the specific threshold configuration.
Furthermore, CLSGen maintains near-perfect parsability ($>0.97$), proving that fine-tuning, unlike zero-shot prompting, successfully enforces structural constraints. This underscores its robustness for clinical data extraction where strict format adherence is critical.

\begin{table}[t]
\small
\resizebox{\columnwidth}{!}{
\centering
\begin{tabular}{lcccccc}
\toprule
 & \multicolumn{1}{c}{Threshold-free}  & \multicolumn{2}{c}{Cohen's Kappa} & \multicolumn{3}{c}{LLM-as-a-Judge}\\
 \cmidrule(lr){2-2} \cmidrule(lr){3-4} \cmidrule(lr){5-7}
Dataset & AUROC-Align & Default & Threshold & RLI & R-L Kappa ($\kappa_{RL}$) & Readability \\
\midrule
LCD 30 days   & 0.9841 & 0.6074 & 0.6187 & 0.0003 & 0.9872 & 0.9996 \\
LCD 90 days   & 0.9597 & 0.6060 & 0.6014 & 0.0018 & 0.9751 & 0.9937 \\
CliniFact     & 0.9891 & 0.8709 & 0.8802 & 0.0482 & 0.8910 & 1.0000 \\
\bottomrule
\end{tabular}
}
\caption{Evaluation of classification-generated text consistency metrics. RLI stands for Rationale-Label Inconsistency, and R-L Kappa  stands for Rationale-Label Kappa metric.}\label{tab:main-aliment}
\end{table}

Table~\ref{tab:main-aliment} summarizes the consistency evaluation between the classification head and the generated text, as well as the internal coherence within the generated output itself. The first group of metrics evaluates the alignment between the classification head’s probability estimations and the verbalized labels from the LM head. Across all datasets, AUROC-Alignment scores range from 0.95 to 0.98, indicating strong agreement between the outputs and suggesting that fine-tuning enables both heads to share a unified internal representation, resulting in consistent predictions. While Cohen’s Kappa also indicates moderate-to-substantial agreement, as a threshold-dependent metric, its limitations are discussed in detail in the Discussion section.

The second group of metrics (LLM-as-a-judge) evaluates the internal consistency between the generated explanation and the verbalized label ($\bar{y}$). Experimental results on RLI show that among parsable outputs, the explanation and label are logically consistent in over 99.8\% of LCD samples and 95\% of CliniFact samples (with an inconsistency rate $\leq 0.049$). Given the exceptionally low RLI and high Rationale-Label Kappa scores, the generated rationale and the predicted label are strongly aligned. These findings provide strong evidence that the provided explanation is closely aligned with the actual classification result.

\section{Discussion}

% \subsection{Threshold-Dependent Metrics}
Threshold-dependent metrics possess inherent drawbacks primarily due to the necessity of arbitrary threshold selection. We identify two major limitations:
First, the optimality of a threshold varies across different metrics. Since the threshold that maximizes one metric (e.g., F1-score) may differ from the one that optimizes another (e.g., Cohen’s Kappa), multi-objective optimization becomes challenging. While adopting a default threshold of 0.5 bypasses this selection dilemma (by being consistently suboptimal across all metrics), it does not resolve the secondary issue regarding performance reliability.
Second, there is a high risk of performance distortion and degradation. Unlike AUROC, which relies on relative sample ranking, threshold-dependent metrics are sensitive to the alignment between the threshold and the data distribution, which is particularly pronounced under class imbalance or small evaluation sets.
Our experimental results on Clinifact (Table~\ref{tab:main-perf}) empirically support this observation. Applying the F1-optimized threshold derived from the development set did not show improvement in F1-score performance compared to using a default threshold of 0.5. This is likely because thresholds tuned on small, skewed splits fail to generalize to the test set. 

One direction for future work is exploring the effect of the algorithm for obtaining explanations to pair with gold labels. Our method used $K=5$ inference runs, and discarded any examples where none of the runs produced an explanation that led to the correct classification.
This means that we are losing some training instances from our (already small) minority class (Appendix~\ref{appendix:data-step-stats} Table~ \ref{tab:data_pipeline_stats}), which may make training more difficult.
On the other hand, it may result in us dropping the noisiest instances. For example, imagine a patient who was extremely stable when discharged but died in a motor vehicle crash in the way home -- there is nothing in the input signal to predict mortality, and if we force the model to re-run until it generates an explanation for why the patient would die, we may be injecting more noise into training.
Exploring this trade-off and qualitatively understanding dropped instances may inform future improvements to the explanation sampling method.

\section{Conclusion}

This study addresses the limitations of Large Language Models (LLMs) in providing reliable probability estimates. While inference-only methods often show sub-optimal performance, conventional fine-tuning approaches frequently trigger ``linguistic collapse.’’ This phenomenon compromises the model's generative capacity, resulting in the loss of verbalized explanations which is a primary advantage of utilizing LLMs. Furthermore, even in cases of successful training, these models often exhibit a lack of alignment between classification outcomes and their supporting justifications.
To overcome these limitations, we proposed CLSGen, a novel framework for binary classification tasks. Experimental evaluations on benchmark datasets demonstrate that CLSGen not only outperforms existing baselines but also achieves superior performance in terms of readability and the alignment between the classification head and the generated text. Consequently, our findings confirm that CLSGen serves as a practical solution to simultaneously enhance discriminative capabilities and preserve the generative capabilities of LLMs.
%Our findings confirm CLSGen as a practical solution for simultaneously enhancing both the discriminative and generative capabilities of LLMs.

\section*{Reproducibility Statement}
To facilitate full reproducibility of our results, we provide comprehensive details across all components of the experimental pipeline:
\begin{itemize}
\item \textbf{Environment:} All experiments were conducted using \texttt{Python 3.10} with \texttt{PyTorch}, \texttt{transformers}, and \texttt{FlashAttention-2}. Training was performed on a multi-GPU Slurm cluster with four NVIDIA A100 (80GB) GPUs in one node. Detailed software versions, hardware specifications, and runtime information are provided in Appendix~\ref{appendix:sec:soft}. 
\item \textbf{Model Configuration and Hyperparameters:} We report complete hyperparameter settings for all methods, including LoRA configurations, optimization details, batch sizes, and training schedules. These are fully specified in Appendix~\ref{appendix:hyperparams}, along with model architecture details for both CLSGen and baseline models. 
\item \textbf{Data Construction:} Synthetic data generation is described in detail in Appendix~\ref{appendix:data_construction}, including the teacher model ($\mathcal{M}_{\text{data}}$), decoding setup, and the use of multiple sampling trials ($K=5$) to ensure label quality and consistency. 
\item \textbf{Model Selection and Evaluation:} We provide a complete description of the model selection protocol, including the definition of the Quality Score and its normalization procedure, as well as threshold tuning strategies and evaluation metrics, in Appendix~\ref{appendix:sec:selection}. 
\item \textbf{Code Availability:} The full implementation, including training scripts, preprocessing pipelines, and prompting templates, will be publicly released upon publication to enable exact replication of our experiments.
\end{itemize}

\section*{Acknowledgments}
Use of LLMs disclosure: The main components of this work utilize LLMs as part of the primary pipeline. LLMs were also used to assist in the preparation of this manuscript for cosmetic purposes (including grammar checking and minor typographical edits). However, the draft was written manually, and the final version was reviewed and edited by the authors. Some code snippets used in the experiments were generated with the assistance of LLMs, but all were reviewed and edited by the authors.

%Funding information hidden for anonymity.
Research reported in this publication was supported by the National Institute Of Mental Health and National Library of Medicine of the National Institutes of Health under Award Numbers R01MH126977 and R01LM012973. The content is solely the responsibility of the authors and does not necessarily
represent the official views of the National Institutes of Health.

\section*{Ethics Statement}
Some of the datasets used in this study are subject to data use agreements due to their inclusion of patient-related records. All experiments were conducted in compliance with these agreements and within a secure, HIPAA-compliant environment.

\bibliography{colm2026_conference}
\bibliographystyle{colm2026_conference}

\newpage
\appendix
\section{Appendix}

\subsection{Statistics of the results}\label{appendix:sec:stats}
In this section, we provide additional statistics for the results reported in the main text, which were omitted due to space constraints.

\begin{table}[h]
\centering
\small
\setlength{\tabcolsep}{5pt}
\begin{tabular}{llccc}
\toprule
Dataset/task & Metric & Verb prob & Label pred & CNN \\
\midrule
\multirow{5}{*}{\shortstack{LCD benchmark \\ 30 days}}
& \textbf{AUROC}  & 0.0124 & N/A    & 0.0015 \\
& \textbf{F1}     & 0.0047 & 0.0123 & 0.0457 \\
& Prec.           & 0.0027 & 0.0097 & 0.0315 \\
& Recall          & 0.0173 & 0.0169 & 0.0354 \\
& Parsable        & 0.0021 & 0.0025 & N/A    \\
\midrule
\multirow{5}{*}{\shortstack{LCD benchmark \\ 90 days}}
& \textbf{AUROC}  & 0.0110 & N/A    & 0.0049 \\
& \textbf{F1}     & 0.0070 & 0.0125 & 0.0061 \\
& Prec.           & 0.0052 & 0.0117 & 0.0183 \\
& Recall          & 0.0109 & 0.0168 & 0.0076 \\
& Parsable        & 0.0046 & 0.0052 & N/A    \\
\midrule
\multirow{5}{*}{Clinifact}
& \textbf{AUROC}  & 0.0101 & N/A    & 0.0040 \\
& \textbf{F1}     & 0.0174 & 0.0112 & 0.0202 \\
& Prec.           & 0.0257 & 0.0118 & 0.0093 \\
& Recall          & 0.0289 & 0.0200 & 0.0428 \\
& Parsable        & 0.0055 & 0.0032 & N/A    \\
\bottomrule
\end{tabular}
\caption{Standard deviations for the performance results in Table~\ref{tab:main-perf}, computed over 10 independent runs. AUROC is unavailable for label prediction, and parsability is not applicable to CNN.}
\label{tab:appendix-std}
\end{table}

\subsection{Implementation Details}\label{appendix:sec:details}

\subsubsection{Software and Training Environment}\label{appendix:sec:soft}
All experiments were implemented using \texttt{Python 3.10.14}. The primary deep learning framework was \texttt{PyTorch v2.4.0}, supported by the \texttt{transformers} library (v4.51.0) and \texttt{FlashAttention-2} \citep{dao2023flashattention}. Computational tasks were executed on a Slurm-configured cluster node equipped with four NVIDIA A100 (80GB) GPUs. The total fine-tuning process took approximately 5 days for the LCD benchmark tasks and 4 hours for CliniFact.

\subsubsection{Model Selection Criteria}\label{appendix:sec:selection}

\paragraph{CLSGen Model Selection:}
In our experimental setup, the primary hyperparameters optimized for reporting test results were the epochs for the best checkpoint and the classification threshold. Other settings were selected empirically without an exhaustive search; a comprehensive list of these values is provided in Section~\ref{appendix:hyperparams}. Checkpoints were saved at the end of each epoch, and the development set was used to select the optimal model for final testing. To determine this "best" epoch of CLSGen model, we defined a Quality Score and selected the checkpoint that maximized this value. Since each metric operates within a different numerical range, we applied z-score normalization to ensure they contributed equally to the final evaluation. The Quality Score is formally defined as:
$$S_{quality} = z(\text{AUROC}_{cls}) + z(\text{AUROC}_{align}) + z(\text{Kappa})$$
where $z(\cdot)$ denotes the z-score function calculated across all checkpoints. 
For threshold-dependent metrics (F1 and Cohen’s kappa), we selected, for each metric, the threshold that maximized its value on the development set and then applied that threshold on the test set. Precision and recall were computed at the F1-optimized threshold.
Notably, during the initial stage of identifying the best epoch via the Quality Score, a default threshold of 0.5 was applied to maintain consistency in metric calculation. For the LCD benchmark, Epoch 5 was identified as the optimal checkpoint for both 30-day and 90-day tasks.

\paragraph{Baseline CNN training:}
For the CNN baseline, both the learning rate and  the best checkpoint (epochs) were selected via grid search. Similar to the CLSGen logic, selection was based on the mean of the AUROC and the $F_1$ score for the minority class on the development set. Final results are reported as the average of 5 independent runs on the test set using these optimized parameters.

For the 30-day task, a learning rate of $5 \times 10^{-5}$ and Epoch 6 were selected; for the 90-day task, a learning rate of $1 \times 10^{-4}$ and Epoch 8 were selected. Final results are reported as the average of 5 independent runs on the test set using these optimized parameters.

\subsubsection{Hyperparameter Specifications}
\label{appendix:hyperparams}

For all LLM-based methods (CLSGen and inference-only baselines),
we used \texttt{Llama-3.1-8B-Instruct} with a maximum input length
of 7,040 tokens and a maximum output length of 384 tokens.
The CNN baseline used the same maximum input length.
Full hyperparameters for CLSGen fine-tuning and the CNN baseline
are listed in Tables~\ref{tab:clsgen_hyperparams}
and~\ref{tab:cnn_hyperparams}, respectively.

\begin{table}[ht]
\centering
\small
\begin{tabular}{ll}
\toprule
\textbf{Hyperparameter} & \textbf{Value} \\
\midrule
Base Model & Llama-3.1-8B-Instruct \\
Max Token Length (Input / Output) & 7{,}040 / 384 \\
Training Epochs & 20 (LCD) / 20 (CliniFact) \\
Batch Size (per GPU) & 1 \\
Gradient Accumulation Steps & 8 \\
Effective Batch Size & 32 \\
Optimizer & AdamW \\
Learning Rate & $2 \times 10^{-5}$ \\
LR Scheduler & Linear warmup, then linear decay \\
Warmup Steps & 1{,}000 (LCD) / 100 (CliniFact) \\
Weight Decay & 0.01 \\
Precision & bfloat16 \\
\midrule
\multicolumn{2}{l}{\textit{LoRA Configuration}} \\
\quad Target Modules & q\_proj, k\_proj, v\_proj, o\_proj \\
\quad Rank ($r$) & 8 \\
\quad $\alpha$ & 16 \\
\quad Dropout & 0.05 \\
\midrule
\multicolumn{2}{l}{\textit{Classification Head}} \\
\quad Hidden Size & 2{,}048 \\
\quad Dropout & 0.1 \\
\bottomrule
\end{tabular}
\caption{Hyperparameters for CLSGen fine-tuning.}\label{tab:clsgen_hyperparams}

\end{table}

\begin{table}[ht]
\centering
\small
\resizebox{\columnwidth}{!}{
\begin{tabular}{ll}
\toprule
\textbf{Configuration} & \textbf{Value} \\
\midrule
\multicolumn{2}{l}{\textit{Grid Search}} \\
\quad Learning Rate & \{5e-4, 1e-4, 5e-5\} \\
\midrule
\multicolumn{2}{l}{\textit{Fixed Hyperparameters}} \\
\quad Batch Size & 16 \\
\quad Max Epochs & 10 \\
\quad Optimizer & AdamW \\
\quad LR Scheduler & Linear warmup $\rightarrow$ Cosine annealing \\
\quad Weight Decay & 0.01 \\
\quad Random Seed & 42 (for dev), 0-4 for test \\
\midrule
\multicolumn{2}{l}{\textit{Architecture}} \\
\quad Pretrained Embeddings & BiomedNLP-BiomedBERT-base-uncased-abstract \citep{pubmedbert}\\%BiomedBERT-base-uncased \\
\quad Num Filters / Kernel Sizes & 512 / \{2, 3, 4, 5\} \\
\quad Projection Dimension & 256 \\
\quad Dropout & 0.5 \\
\quad Pooling & Global Max Pooling \\
\bottomrule
\end{tabular}
}
\caption{Hyperparameters and architecture for the CNN baseline.}\label{tab:cnn_hyperparams}

\end{table}

\subsubsection{Data Construction Details}
\label{appendix:data_construction}

For data construction, we used \texttt{DeepSeek-R1-Distill-Llama-8B} as $\mathcal{M}_{\text{data}}$,  with maximum input and output lengths of 8,196 and 2,048 tokens, respectively. The longer output length accommodates the model's chain-of-thought reasoning tokens (enclosed in \texttt{<think>} tags), which precede the final output. Since data construction is an inference-only process, this extended output length is feasible without the memory constraints associated with fine-tuning. We performed up to $K=5$ generation trials per instance to ensure high-quality synthetic labels and justifications. 

We selected a model with strong reasoning capabilities as $\mathcal{M}_{\text{data}}$ to maximize the quality of generated explanations and labels. Importantly, $\mathcal{M}_{\text{base}}$ (\texttt{Llama-3.1-8B-Instruct}) is a different model from $\mathcal{M}_{\text{data}}$, demonstrating that CLSGen does not require the data-generating model and the fine-tuned model to be identical. This separation highlights the generalizability of the framework: a model with stronger reasoning ability produces high-quality training signal, which is then distilled into the target model through fine-tuning.

%\subsection{Thresholds}

\subsection{Data construction step statistics}\label{appendix:data-step-stats}

Table~\ref{tab:data_pipeline_stats} shows the statistics regarding the data construction step.

\begin{table}[h!]
\small
\centering
\begin{tabular}{clccc}
\toprule
\textbf{Dataset} & \textbf{Status} & \textbf{False} & \textbf{True} & \textbf{True / Total} \\
\midrule
\multirow{3}{*}{\shortstack{LCD-bench 30 days\\ (Dev)}} 
& Original dataset        & 7{,}203 & 302 & 4.0\% \\
& Successfully generated  & 7{,}102 & 164 & 2.3\% \\
& Failed to generate      & 101     & 138 & 57.7\% \\
\midrule
\multirow{3}{*}{\shortstack{LCD-bench 90 days\\ (Dev)}} 
& Original dataset        & 7{,}203 & 302 & 4.0\% \\
& Successfully generated  & 6{,}950 & 198 & 2.8\% \\
& Failed to generate      & 253     & 104 & 29.1\% \\
\midrule
\multirow{3}{*}{\shortstack{CliniFact\\(Dev)}} 
& Original dataset        & 233 & 83 & 26.3\% \\
& Successfully generated  & 219 & 65 & 22.9\% \\
& Failed to generate      & 14  & 18 & 56.3\% \\
\bottomrule
\end{tabular}
\caption{Development dataset data processing pipeline statistics.}
\label{tab:data_pipeline_stats}
\end{table}

\subsection{Example prompts}\label{appendix:sec:prompt}

In this section, we present the prompts used in our experiments.

\paragraph{Main training experiments}

The prompts used during CLSGen training are structured as follows. The input part is constructed by concatenating \texttt{system prompt}, \texttt{user prompt (document)}, \texttt{question}, and \texttt{generation prefix} in this order. Notably, generation prefix is placed at the very end of the input. The final token of this sequence is fed into the LLM, and the resulting hidden representation is used as the input to the CLS head.

The target part consists of the \texttt{response template}. Importantly, an End of Generation (EOG) token is appended at the end to indicate the termination of generation.
Below is an example used for training on the LCD benchmark dataset.

\begin{verbatim}
system_prompt = f"You are a clinical decision support assistant. Analyze the 
clinical document provided and classify the patient's out-of-hospital mortality 
within {task_days} days as either 0:alive or 1:death. First provide your clinical 
reasoning, then state your final classification."

user_prompt = "Clinical Document:\n\n{document}" 

question = f"Based on the clinical document below, reason about the patient's 
likely outcome and classify their out-of-hospital mortality within {task_days} 
days. Answer with exactly one of: 0:alive or 1:death."

generation_prefix = "Reasoning: " 
# Not the reasoning prompt used within <think> </think>

response_template = "{explanation}\n\nClassification: {label}\n\nEOG"
\end{verbatim}

\paragraph{Label prediction}
For label prediction, we used the prompt for the main training experiments, except for the response template. 

\paragraph{Verbalized probability}
\begin{verbatim}
SYSTEM_PROMPT = (
    "You are a clinical decision support assistant. "
    "Analyze the clinical document provided and estimate the probability of "
    "out-of-hospital mortality within {task_days} days."
)

QUESTION = (
    "Based on the clinical document below, what is the probability (0 to 100) "
    "that this patient will die out-of-hospital within {task_days} days?\n\n"
    "Briefly reason about the key clinical findings, then state your answer "
    "in the format: Probability: <integer>"
)
\end{verbatim}

\subsubsection{LLM-as-a-judge}

We measure the quality of the text and the inconsistency between the explanation and the verbalized label prediction within the same output generated by the CLSGen model. To this end, we adopt an LLM-as-a-judge setup. The judge model is prompted with the following system instruction and evaluation template:

\paragraph{Inconsistency measure}

\begin{verbatim}
JUDGE_SYSTEM_PROMPT = """You are a text analyst. You will be shown a piece of 
reasoning written by another model.
Read the surface meaning of the text and determine what conclusion it implies.
Do not use external knowledge — judge solely based on what the text says.
Output ONLY the classification label. Do not explain or add any other text."""

JUDGE_TEMPLATE_CLINIFACT = """Read the following reasoning and determine whether 
it implies the claim is supported or not.

REASONING:
\"\"\"
{explanation}
\"\"\"

Reply with ONLY one of the two lines below — no explanation, no extra text:
Classification: TRUE
Classification: FALSE"""
\end{verbatim}

\paragraph{Readability measure}

\begin{verbatim}
JUDGE_SYSTEM_PROMPT = """You are a text quality evaluator. You will be shown a 
piece of text written by a language model.
Judge whether the text is readable and coherent.
Output ONLY the verdict label. Do not explain or add any other text."""

JUDGE_TEMPLATE = """Read the following text and decide whether it is readable.

Mark it as UNREADABLE if it contains ANY of the following:
- Made-up or nonsensical words
- Unexpected foreign characters or scripts (e.g. random Chinese, Arabic, or other
non-English characters)
- A sentence that cuts off abruptly or is clearly incomplete
- Content that is largely unintelligible

Otherwise mark it as READABLE.

TEXT:
\"\"\"
{text}
\"\"\"

Reply with ONLY one of the two lines below — no explanation, no extra text:
Readability: READABLE
Readability: UNREADABLE"""
\end{verbatim}

\end{document}